# Linear Algebra Approach to Separable Bayesian Networks

**Chalee Asavathiratham**
Pivotal Systems Corporation
4637 Chabot Drive
Pleasanton, CA 94588
chalee@alum.mit.edu

## Abstract

Separable Bayesian Networks, or the Influence Model, are dynamic Bayesian Networks in which the conditional probability distribution can be separated into a function of only the marginal distribution of a node's parents, instead of the joint distributions. We describe the connection between an arbitrary Conditional Probability Table (CPT) and separable systems using linear algebra. We give an alternate proof to [Pfeffer00] on the equivalence of sufficiency and separability. We present a computational method for testing whether a given CPT is separable.

## 1 Introduction

Separable Bayesian Networks, or the Influence Model, are dynamic Bayesian Networks in which the conditional probability distribution can be separated into a function of only the marginal distribution of a node's neighbors, instead of the joint distributions. In terms of modeling, separable networks has rendered possible siginificant reduction in complexity, as the state space is only linear in the number of variables on the network, in contrast to a typical state space which is exponential.

We describe the connection between an arbitrary Conditional Probability Table (CPT) and separable systems using linear algebra. We give an alternate proof to [Pfeffer00] on the equivalence of sufficiency and separability. We present a computational method for testing whether a given CPT is separable.

## 2 Results

### 2.1 Probability Mass Functions

Suppose $X$, $Y$ and $Z$ are three random variables and let the size of their sample spaces be some finite integers $m_x$, $m_y$ and $m_y$ respectively. Similar to [Pfeffer00], we denote the space of *Probability Mass Functions* (PMF) over the joint random variables $(X,Y)$ as $\Delta^{XY}$, i.e.,

$$\Delta^{XY} \triangleq \{\, \mathbf{q} \in \mathcal{R}^{m_x m_y} \mid \mathbf{q} \geq 0,\ \mathbf{q}'\mathbf{1}_{m_x m_y} = 1 \},$$

where $\mathbf{1}_{m_x m_y}$ is the all-ones column vector of length $(m_x m_y)$, and $\mathbf{q} \geq 0$ means each entry of the column vector $\mathbf{q}$ is non-negative. When there is no ambiguity, we will drop the subscript from $\mathbf{1}$.

### 2.2 Event Matrix

We now introduce a particular class of matrices which would be useful for our discussion later. These matrices have been introduced in [Asavathiratham00] (Sec. 5.2.1) and is instrumental in some of the key theoretical results in the thesis.

Let $\mathbf{e}_i = [\,0 \cdots 0\ 1\ 0 \cdots 0\,]'$ be the vector in which the single 1-entry appears in the $i$th position. For random variable $X$, we can represent its $i$th outcome as $\mathbf{e}_i \in \mathcal{R}^{m_x}$. Stacking these vectors into a matrix, we get the *event matrix* $B_X$ – in this case, the $m_x$-by-$m_x$ identity matrix – whose rows form a sample space for $X$.

For joint variables $(X,Y)$, let us define the event matrix as

$$B_{XY} \triangleq [\,I_{m_x} \otimes \mathbf{1}_{m_y} \mid \mathbf{1}_{m_x} \otimes I_{m_y}\,],$$

where $J \otimes K$ denotes the Kronecker product of matrices

$J$ and $K$.[1] For example, if $m_x = 2$ and $m_y = 3$, then

$$B_{XY} = \begin{bmatrix} 1 & 0 & | & 1 & 0 & 0 \\ 1 & 0 & | & 0 & 1 & 0 \\ 1 & 0 & | & 0 & 0 & 1 \\ 0 & 1 & | & 1 & 0 & 0 \\ 0 & 1 & | & 0 & 1 & 0 \\ 0 & 1 & | & 0 & 0 & 1 \end{bmatrix}.$$

For convenience, let us define $B \triangleq B_{XY}$. It can be seen that each row of $B$ is an outcome of the joint random variable of $(X,Y)$. Moreover, the ordering of the row $B$ also provides for us an index for elements in the sample space of $(X,Y)$. Note that

$$B \begin{bmatrix} \mathbf{1}_{m_x} \\ 0 \end{bmatrix} = B \begin{bmatrix} 0 \\ \mathbf{1}_{m_y} \end{bmatrix} = \mathbf{1}_{m_x m_y}. \quad (1)$$

We shall also list some properties of $B$ that will be used later on in the paper.

**Theorem 1**

$$rank(B) = m_x + m_y - 1$$

**Proof:** See Theorem 5.8 in [Asavathiratham00]. □

Let $\mathcal{N}(K)$ and $\mathcal{R}(K)$ be the null space and the subspace of some matrix $K$.

**Corollary 2**

$$\mathcal{N}(B) = \mathcal{R}\left( \begin{bmatrix} \mathbf{1}_{m_x} \\ -\mathbf{1}_{m_y} \end{bmatrix} \right)$$

**Proof:** See Corollary 5.10 in [Asavathiratham00]. □

Note that for a given $\mathbf{q} \in \Delta^{XY}$, the product

$$\mathbf{q}'B = [\, \mathbf{q}'_X \; \mathbf{q}'_Y \,],$$

where $\mathbf{q}_X$ and $\mathbf{q}_Y$ are marginal of $\mathbf{q}$ over $X$ and $Y$ respectively. That is, multiplication by $B$ gives us the marginal distribution on $X$ and $Y$ from $\mathbf{q}$. As such, we define the *marginalization matrices* as follows

$$B_{XY,X} \triangleq B \begin{bmatrix} B'_X \\ 0 \end{bmatrix} \quad (2)$$

$$B_{XY,Y} \triangleq B \begin{bmatrix} 0 \\ B'_Y \end{bmatrix}, \quad (3)$$

so that $\mathbf{q}'_X = \mathbf{q}'B_{XY,X}$ and $\mathbf{q}'_Y = \mathbf{q}'B_{XY,Y}$. These marginilization matrices shall be used below.

---
[1] For example, if $J$ is a 2-by-2 matrix with entries given by $j_{11}, \ldots, j_{22}$ and if $K$ is some arbitrary matrix, then

$$J \otimes K = \begin{bmatrix} j_{11}K & j_{12}K \\ j_{21}K & j_{22}K \end{bmatrix}.$$

## 2.3 Sufficiency

For a given conditional PMF $P(Z|XY)$, we let $C_{P(Z|XY)}$, or simply $C$, denote its *Conditional Probability Table* (CPT). That is, $C$ is a $(m_x m_y)$-by-$m_z$ matrix such that each row of $C$ corresponds to an outcome of $(X,Y)$, sorted in the order defined by $B$. Specifically, let the $(i,j)$th entry of $C$ be defined as

$$[C]_{ij} \triangleq P(\, Z = j \mid (X,Y) \text{ is } i\text{th row of } B\,)$$

It then follows that $C$ has nonnegative entries and

$$C \mathbf{1}_{m_z} = \mathbf{1}_{m_x m_y}.$$

In [Pfeffer00], the function $\Phi^P : \Delta^{XY} \to \Delta^Z$ was defined as $\Phi^P(\mathbf{q}) \triangleq \sum_{xy} \mathbf{q}(xy) P(Z|xy)$. Expressing this definition in a matrix form, we get

$$\Phi^P(\mathbf{q}) \triangleq \mathbf{q}'C.$$

Because $\Phi^P$, $P(Z|XY)$ and $C$ completely describe one another, we will focus mainly on $C$.

In general, obtaining a marginal on $Z$ requires that we know the full joint distribution on $(X,Y)$. Below we define a special condition such that the marginals on $X$ and $Y$ alone can uniquely determine the marginal on $Z$.

**Definition 1** $C$ *is* sufficient *if for any* $\mathbf{q}_1, \mathbf{q}_2 \in \Delta^{XY}$ *such that* $\mathbf{q}'_1 B = \mathbf{q}'_2 B$, *then* $\mathbf{q}'_1 C = \mathbf{q}'_2 C$.

**Theorem 3** $C$ *is sufficient if and only if* $\mathcal{N}(B') \subset \mathcal{N}(C')$

**Proof:** First suppose $C$ is sufficient. Take any $\tilde{\mathbf{q}} \in \mathcal{N}(B')$ so that $\tilde{\mathbf{q}}'B = 0$.

If $\tilde{\mathbf{q}} \geq 0$, then it must follow that $\tilde{\mathbf{q}} = 0$. Otherwise, if $\tilde{\mathbf{q}}$ has any positive entry, $\tilde{\mathbf{q}}'B$ would have at least some positive entries, because every row of $B$ has some positive entries. This would contradict the fact $\tilde{\mathbf{q}}'B = 0$.

Then $\tilde{\mathbf{q}}$ must have at least one negative entry. This implies that it must also have at least one positive entry, because

$$\tilde{\mathbf{q}}'\mathbf{1} = \tilde{\mathbf{q}}'B \begin{bmatrix} \mathbf{1}_{m_x} \\ 0 \end{bmatrix} = 0. \quad (4)$$

where the second equality follows from (1). We can, therefore, separate the positive entries from the negative ones to obtain

$$\tilde{\mathbf{q}} = \mathbf{q}_+ - \mathbf{q}_-,$$

where $\mathbf{q}_+$ and $\mathbf{q}_-$ are the absolute values of the positive and negative entries of $\tilde{\mathbf{q}}$ respectively. Let $d \triangleq$

$\mathbf{q}'_+\mathbf{1} = \mathbf{q}'_-\mathbf{1}$, where the second equality follows from (4). Because $\mathbf{q}_+$ has some positive entry, $d > 0$. Define $\tilde{\mathbf{q}}_+ = \frac{1}{d}\mathbf{q}_+$ and $\tilde{\mathbf{q}}_- = \frac{1}{d}\mathbf{q}_-$. It can then be shown that $\tilde{\mathbf{q}}_+$ and $\tilde{\mathbf{q}}_-$ are both valid PMFs. Therefore,

$$\begin{aligned}
\tilde{\mathbf{q}}'B = 0 &\leftrightarrow d(\tilde{\mathbf{q}}_+ - \tilde{\mathbf{q}}_-)'B = 0 \\
&\leftrightarrow \tilde{\mathbf{q}}'_+ B = \tilde{\mathbf{q}}'_- B \\
&\leftrightarrow \tilde{\mathbf{q}}'_+ C = \tilde{\mathbf{q}}'_- C \quad (5)\\
&\leftrightarrow \tilde{\mathbf{q}}'C = 0
\end{aligned}$$

where (5) follows from sufficiency of $C$. This completes the forward part of the proof. The converse of the proof is straightforward. □

Let $\mathcal{R}(A)$ denote the subspace spanned by the matrix $A$.

**Corollary 4** $C$ is sufficient if and only if $\mathcal{R}(C) \subset \mathcal{R}(B)$.

**Proof:** This follows from the fact that $\mathcal{N}(B') \subset \mathcal{N}(C')$ is equivalent to $\mathcal{R}(C) \subset \mathcal{R}(B)$. □

## 2.4 Separability

First we state the definition of separability as given in [Pfeffer00].

**Definition 2** $P(Z|XY)$ is separable *if there exists condition distributions $P_X(Z|X)$, $P_Y(Z|Y)$ and a constant $\gamma \in [0,1]$ such that $P(Z|XY) = \gamma P_X(Z|X) + (1-\gamma)P_Y(Z|Y)$.*

Separability is the key to the reduction in complexity. To see this, consider, for example, the task of listing $P(Z|XY)$ for all possible values of $X, Y$ and $Z$. In general, we would require a table whose number of entries is on the order of $O(m_x m_y m_z)$. Through separability, we only need to list the conditional probabilities $P(Z|X)$ and $P(Z|Y)$ separately, and compute $P(Z|XY)$ upon needed. This would result in a table on the order of $O(m_z m_x + m_z m_y)$. The reduction would be even greater as we extend this to multiple variables such as $P(Z|X_1 X_2 \cdots X_n)$ later on.

**Proposition 1** $P(Z|XY)$ *is separable if and only if $C$ can be expressed as*

$$C = \gamma B_{XY,X} C_X + (1-\gamma) B_{XY,Y} C_Y \quad (6)$$

*where $C_X$ and $C_Y$ are $m_x$-by-$m_z$ and $m_y$-by-$m_z$ non-negative matrices with every row summing to 1.*

To see why this is the case, one only has to recognize that that the rows of $C_X$ and $C_Y$ are respectively the conditional PMFs $P_X(Z|X)$ and $P_Y(Z|Y)$ expressed in a matrix form, with one row for each different outcome of $X$ and $Y$. If we multiply (6) by some $\mathbf{q} \in \Delta^{XY}$, the left hand-side $\mathbf{q}'C$ would be a marginal distribution $P(Z)$. On the right-hand side, the first-term would be

$$\mathbf{q}'B_{XY,X}C_X = \mathbf{q}'_X C_X,$$

which is $P_X(Z|X)$ in vector form. Similarly, the second-term on the right-hand side would be $P_Y(Z|Y)$ as desired.

**Example 1** Let $(m_x, m_y, m_z) = (2, 3, 2)$ and let

$$C = \begin{bmatrix} .65 & .35 \\ .35 & .65 \\ .20 & .80 \\ .60 & .40 \\ .30 & .70 \\ .15 & .85 \end{bmatrix}.$$

One way to factorize $C$ into the form in (6) is to as follows.

$$\gamma = 0.5, \quad C_X = \begin{bmatrix} .3 & .7 \\ .2 & .8 \end{bmatrix}, \quad C_Y = \begin{bmatrix} 1 & 0 \\ .4 & .6 \\ .1 & .9 \end{bmatrix}.$$

Since the factorization is not unique, another choice would be

$$\gamma = 0.5, \quad C_X = \begin{bmatrix} .4 & .6 \\ .3 & .7 \end{bmatrix}, \quad C_Y = \begin{bmatrix} .9 & .1 \\ .3 & .7 \\ 0 & 1 \end{bmatrix}.$$

□

The following result has been first shown in [Pfeffer00]. The proof we are presenting below is different in that it is based on linear algebra. By establishing this connection, we hope to apply some of the results from matrix analysis to simplify some theoretical and computational problems in the Bayesian Network area.

**Theorem 5** $C$ is sufficient if and only if $P(Z|XY)$ is separable.

**Proof:** First suppose that $C$ is sufficient. By Corollary 4, $C = BF$ for some $(m_x + m_y)$-by-$m_z$ matrix $F$. Because each row of $C$ is a valid PMF, $C\mathbf{1}_{m_z} = \mathbf{1}_{m_x m_y}$, which means

$$BF\mathbf{1}_{m_z} = \mathbf{1}_{m_x m_y}. \quad (7)$$

We would like factorize $F$ in (7) to show its separability. From (1), we know that

$$B \cdot \frac{1}{2}\begin{bmatrix} \mathbf{1}_{m_x} \\ \mathbf{1}_{m_y} \end{bmatrix} = \frac{1}{2}(\mathbf{1}_{m_x m_y} + \mathbf{1}_{m_x m_y}) = \mathbf{1}_{m_x m_y}. \quad (8)$$

Combining (8) and Corollary 2, we can express the term $F\mathbf{1}$ as a sum of particular and homogeneous solution, i.e.,

$$F\mathbf{1}_{m_z} = \frac{1}{2}\begin{bmatrix} \mathbf{1}_{m_x} \\ \mathbf{1}_{m_y} \end{bmatrix} + \delta \begin{bmatrix} \mathbf{1}_{m_x} \\ -\mathbf{1}_{m_y} \end{bmatrix} \quad (9)$$

where $\delta$ is some constant.

*Case A:* If $F \geq 0$, then (9) implies that $-\frac{1}{2} \leq \delta \leq \frac{1}{2}$. For the special cases when $-\frac{1}{2} < \delta < \frac{1}{2}$, we let $\gamma = \frac{1}{2} + \delta$ and

$$C_X = \frac{1}{\frac{1}{2}+\delta} [I_{m_x} \; 0] F$$

$$C_Y = \frac{1}{\frac{1}{2}-\delta} [0 \; I_{m_y}] F$$

If, however, $\delta = \frac{1}{2}$, then we let $\gamma = 1$ and $C_X = [I_{m_x} \; 0] F$, and $C_Y$ can be any CPT. Similarly, when $\delta = -\frac{1}{2}$, we let $C_Y = [0 \; I_{m_y}] F$, and $\gamma = 0$.

It can easily verified that $C_X$ and $C_Y$ as defined are valid CPTs, and that $\gamma$ would lie in $[0,1]$. Substituting these values into (6), we thus have proved that $P(Z|XY)$ is separable.

*Case B:* If $F$ has some negative entries, we can convert it to a non-negative matrix one column at a time as follows. Let $\mathbf{f}$ be a column that contains at least one negative entry. Let us partition it into blocks of $m_x$ and $m_y$ entries so that $\mathbf{f}' = [\mathbf{f}'_x \; \mathbf{f}'_y]$. Each row of $B\mathbf{f}$ can be written as

$$(\mathbf{f}_x)_i + (\mathbf{f}_y)_j \qquad (10)$$

where $(\mathbf{f}_x)_i$ and $(\mathbf{f}_y)_j$ represents some $i$th and $j$th entries of $\mathbf{f}_x$ and $\mathbf{f}_y$ respectively. Because $BF = C \geq 0$, at most one of the terms in (10) can be negative. Indeed, if $\mathbf{f}_x$ has a negative entry, there can be no negative value in $\mathbf{f}_y$ at all, and vice versa. Assume without loss of generality that $\mathbf{f}_y \geq 0$, and $\mathbf{f}_x$ contains some negative entry. Let $i^* \triangleq \arg\min_i (\mathbf{f}_x)_i$, and $j^* \triangleq \arg\min_j (\mathbf{f}_y)_j$. Because $B\mathbf{f} \geq 0$, $(\mathbf{f}_y)_{j^*} \geq -(\mathbf{f}_x)_{i^*} > 0$. Then replace $\mathbf{f}$ with

$$\tilde{\mathbf{f}} \triangleq \mathbf{f} + \alpha \begin{bmatrix} \mathbf{1}_{m_x} \\ -\mathbf{1}_{m_y} \end{bmatrix},$$

where $-(\mathbf{f}_x)_{i^*} \leq \alpha \leq (\mathbf{f}_y)_{j^*}$. By construction, $\tilde{\mathbf{f}}$ would have no negative entry. Furthermore, because of Corollary 2, $B\mathbf{f} = B\tilde{\mathbf{f}}$. We can apply this to all other columns of $F$ that has a negative entry until we arrive at $\tilde{F} \geq 0$. Then we can apply Case A to obtain $C_X$ and $C_Y$. This thus proves that $P(Z|XY)$ is separable.

On the other hand, if $P(Z|XY)$ is separable, then we can factorize $C$ according to Proposition 1. By (2)-(3), $B_{XY,X}, B_{XY,Y} \in \mathcal{R}(B)$. Thus, $C \in \mathcal{R}(B)$. By Corollary 4, $C$ is sufficient. □

### 2.5 Extension to Multiple Variables

The results above can be extended to the multiple-variable case such as $P(Z|X_1 \cdots X_n)$.

Let $m_1, \ldots, m_n$ be the size of the sample spaces of $X_1, \ldots, X_n$ respectively. We construct a sequence of matrices $\{B_i\}$ from $i=1$ to $i=n$ through a recursive procedure as follows:

$$B_1 \triangleq I_{m_1}$$
$$B_i \triangleq [\, B_{i-1} \otimes \mathbf{1}_{m_i} \mid \mathbf{1}_{\mu_{i-1}} \otimes I_{m_i} \,] \qquad (11)$$

where $\mu_i \triangleq \prod_{j=1}^i m_j$. The *event matrix* and, will be denoted by $B \triangleq B_n$. Some properties of $B$ are presented again for the general case.

**Theorem 6** *For $1 \leq i \leq n$,*

$$\text{rank}(B_i) = \left(\sum_{k=1}^{i} m_k\right) - i + 1$$

**Proof:** See Theorem 5.8 in [Asavathiratham00]. □

**Theorem 7**

$$\mathcal{N}(B) = \left\{ \mathbf{v} \mid \mathbf{v} = \begin{bmatrix} a_1 \mathbf{1}_{m_1} \\ \vdots \\ a_n \mathbf{1}_{m_n} \end{bmatrix} \text{ with } \sum_i a_i = 0 \right\}$$

**Proof:** See Corollary 5.10 in [Asavathiratham00]. □

**Definition 3 (General Sufficient Condition)** *$C$ is sufficient if for any $\mathbf{q}_1, \mathbf{q}_2 \in \Delta^{X_1 \cdots X_n}$ such that $\mathbf{q}'_1 B = \mathbf{q}'_2 B$, then $\mathbf{q}'_1 C = \mathbf{q}'_2 C$.*

Thereom 3 and Corollary 4 and their proofs still hold in the general case.

**Definition 4 (General Sepability Condition)** *$P(Z|X_1 \cdots X_n)$ is separable if there exist a set of distributions $\{P_i(Z|X_1 \cdots X_n)\}$ and non-negative constants $\{\gamma_i\}$ such that $\sum_i \gamma_i = 1$ and*

$$P(Z|X_1 \cdots X_n) = \sum_{i=1}^n \gamma_i P_i(Z|X_i).$$

Let $\hat{B}_i = B \cdot (\mathbf{e}_i \otimes I_{m_i})$, i.e., $\hat{B}_i$ extracts the columns corresponding to the $i$th variable from $B$.

**Proposition 2** *$P(Z|X_1 \cdots X_n)$ is separable if and only if $C$ can be written as*

$$C = \sum_{i=1}^n \gamma_i \hat{B}_i C_i \qquad (12)$$

*where each $C_i$ is an $m_i$-by-$m_z$ nonnegative matrix in which every row sums to 1, and the coefficients $\{\gamma_i\}$ are non-negative and sum to 1.*

**Theorem 8** $C$ is sufficient if and only if $P(Z|X_1 \cdots X_n)$ is separable.

**Sketch of Proof:** The proof would proceed similarly as the one in Theorem 5, except that (9) should be written as

$$F\mathbf{1} = \frac{1}{n}\begin{bmatrix} \mathbf{1}_{m_1} \\ \vdots \\ \mathbf{1}_{m_n} \end{bmatrix} + \begin{bmatrix} a_1\mathbf{1}_{m_1} \\ \vdots \\ a_n\mathbf{1}_{m_n} \end{bmatrix},$$

in which $\sum_i a_i = 0$ from Theorem 7.

*Case A:* If $F \geq 0$, we apply the same arguments and let $C_i = \frac{1}{\frac{1}{n}+a_i}\begin{bmatrix} 0 \cdots I_{m_i} \cdots 0 \end{bmatrix} F$.

*Case B:* If $F$ has some negative entry, we would again add some columns to its to make it non-negative. Here we partition $\mathbf{f}$ into $n$ blocks according to $\{m_i\}$. Then there must be some row of $B\mathbf{f}$ that is equal to

$$\sum_{i=1}^{n} \min(\mathbf{f}_i), \qquad (13)$$

i.e., some row of $B$ would happen to select the minimal entry of every block $\{\mathbf{f}_i\}$ into the sum. Let us group all the indices that contribute positive (or zero) value in (13) into $A_+$, and let the rest be $A_-$. Now, for all $i \in A_-$, let $\alpha_i = |\min(\mathbf{f}_i)|$. For $i \in A_+$, let $\alpha_i$ be some negative coefficients chosen in any way such that $\alpha_i + \min(\mathbf{f}_i) \geq 0$ and $\sum_{i \in A_+} \alpha_i = -\sum_{i \in A_-} \alpha_i$. These $\{\alpha_i\}$ must exist[2] because

$$\sum_{i \in A_+} \min(\mathbf{f}_i) \geq -\sum_{i \in A_-} \min(\mathbf{f}_-),$$

which follows from the fact that $B\mathbf{f} \geq 0$. Thus, by replacing $\mathbf{f}$ with

$$\tilde{\mathbf{f}} = \mathbf{f} + \begin{bmatrix} \alpha_1 \mathbf{1}_{m_1} \\ \vdots \\ \alpha_n \mathbf{1}_{m_n} \end{bmatrix},$$

we would have a non-negative column as desired. The rest of the proof is the same. □

### 2.6 Test for Separability

Given a CPT matrix $C$ we can test if it is separable by simply verifying that it lies in the subspace of $B$. One way to do this is to test whether the orthogonal projection $C$ onto $\mathcal{R}(B)$ is equal to itself. If $C \notin \mathcal{R}(B)$, then the projection would only be the least-square approximation.

To achieve this, we would need a basis for $\mathcal{R}(B)$. Unfortunately, $B$ itself does not have full column rank according to Theorem 6 and thus cannot be directly used to create an orthogonal projection. We propose the following slightly modified matrix $A$, which is defined in the following recursive way:

$$A_1 \triangleq I_{m_1}$$
$$A_i \triangleq [\, A_{i-1} \otimes \mathbf{1}_{m_i} \mid \mathbf{1}_{\mu_{i-1}} \otimes \tilde{I}_{m_i} \,]$$

where $\mu_i \triangleq \prod_{j=1}^{i} m_j$, and $\tilde{I}_m$ is the matrix consisting of the first $m-1$ columns of the $m$-by-$m$ identity matrix. The resulting $A \triangleq A_n$ would be a matrix with $\mu_n$ rows and $(\sum_i m_i) - n + 1$ columns. By inspection, one can see $A$ can be obtained from $B$ by dropping certain $n-1$ columns. This means that $\mathcal{R}(A) \subset \mathcal{R}(B)$.

---
[2]For example, a greedy algorithm would work.

**Theorem 9** *The columns of $A$ are linearly independent and spans $\mathcal{R}(B)$.*

**Proof:** For $i = 1$ the claim is trivially true. Assume that the assertion is true up to some index $i - 1$. Because $B_{(i-1)} \in \mathcal{R}(A_{(i-1)})$, there exists some matrix $K$ such that $B_{(i-1)} = A_{(i-1)}K$. Thus, by the Kronecker product property, $B_{i-1} \otimes \mathbf{1}_{m_i} = (A_{(i-1)} \otimes \mathbf{1}_{m_i})K$. Define

$$\tilde{B}_i \triangleq [\, B_{(i-1)} \otimes \mathbf{1}_{m_i} \mid \mathbf{1}_{\mu_{i-1}} \otimes \tilde{I}_{m_i} \,] \qquad (14)$$

From Lemma B.1 in Appendix B of [Asavathiratham00], we have that

$$\begin{aligned} rank(\tilde{B}_i) &= rank(B_{(i-1)}) + m_i - 1 \\ &= rank(A_{(i-1)}) + m_i - 1 \quad (15) \\ &= rank(A_i). \quad (16) \end{aligned}$$

Eq. (15) comes from assertion of the proof up to $i-1$, and (16) is due to the fact that $m_i - 1$ new columns are added in going from $A_{(i-1)}$ to $A_i$. □ [3]

Define the *orthogonal projection* of $C$ as

$$\mathcal{P}(C) \triangleq A(A'A)^{-1}A'C. \qquad (17)$$

Note that the inverse $(A'A)^{-1}$ must exist because $A$ has full column rank.

**Theorem 10 (Separability Test)** *$C$ is separable if and only if $\mathcal{P}(C) = C$.*

**Proof:** This follows from standard linear algebra theory and from Corollary 4 and Theorem 8. □

### 2.7 Connection to the Influence Model

The Influence Model was first introduced in [Asavathiratham00] and succinctly described in [Asavathiratham01]. Originally intended to explain cascading phenomena in power systems, the model has been applied in applications such as social interaction, viral marketing networks, information diffusion, and distributed control.

---
[3]Can we find another basis that is both orthonormal and recursive?

Recall that an Influence Model is defined by an $n$-by-$n$ stochastic matrix $D$ called the *influence matrix* and a set of matrices $\{A_{ij}\}$ for all $i, j = 1, \ldots, n$ such that every $A_{ij} \geq 0$ and $A_{ij}\mathbf{1} = \mathbf{1}$. Each site $i$ (node $i$) at time $k$ has a status that is represented by

$$\mathbf{s}_i[k] = [\ 0 \cdots 1 \cdots 0\ ]'$$

where the position of the 1-entry represents the its current status. For each time step, given the states $\{\mathbf{s}_i[k]\}$ the next-state PMF, or the posterior belief, at time $k+1$ state $i$ is given by

$$\mathbf{p}'_i[k+1] = d_{i1}\mathbf{s}'_1[k]A_{i1} + \cdots + d_{in}\mathbf{s}'_n[k]A_{in}.$$

The state at time $k+1$ would be realized randomly according to $\mathbf{p}_i[k+1]$.

It turns out that a DBN in which all CPT's are separable is an influence model. To make the relation explicit, one can see that the entries in each row of $D$ is equivalent to a set of $\{\gamma_i\}$ in (12), and $\{A_{ij}\}$ are the $C_i$'s.

Approximating a given DBN with an Influence Model should allow us computational savings in various ways. For example, the Influence Model allows us to compute the marginal distribution of a node at any given time efficiently. It also provides for a simple way for analyzing the recurrent classes based on reduced-order graph structure. Absorption probabilities of each recurrent classes can also be efficiently computed.

## 3 Open Issues

### 3.1 Approximating General DBN with a Separable System

So far, we have only focused on the CPT of a given variable $Z$, but the idea can be extended to a network of random variables. That is, if we are given a Dynamic Bayesian Network in which every node has an arbitrary CPT based on the joint distribution of its neighbors (including possibly itself), then we can appoximate it with a separable version. What would then be the optimal way to achieve this?

One possible procedure is to work with the CPT of each node separately. That is, one approximates the CPT table at each node with a separable version that is optimal in some sense. Once we have a separable CPT, we can factorize it into smaller CPT tables according to (12).

What should be the optimality criterion in the approximation? Although $\mathcal{P}(C)$ is optimal in the least-square sense, i.e.,

$$\mathcal{P}(C) = \arg \min_{X \in \mathcal{R}(B)} ||X - C||^2, \qquad (18)$$

there might be other objective function that serves us better such as the Kullback-Leibler distance between the original and the approximated distributions.

### 3.2 Learning and Inference on Separable Systems

Separable systems may offer a significant advantage over general DBNs in terms of speed in parameter learning and state inference. Because the number of parameters for such systems is usually much smaller than that for a general DBN, the parameter learning should require much less data and behave much more stably. This benefit has been the key to the application demonstrated in [Basu01].

For a general DBN, the task of updating the posterior distribution still requires a computation that is exponential in number of variables in the system, for instance the forward-backward algorithm. Given that separable BNs are defined so that the marginal distributions are sufficient to predict themselves in the future, it seems plausible that there exists some method for the inference tasks with reduced computation.

**Acknowledgements**

The author wishes to thank Prof. Benjamin Van Roy (Stanford), Prof. George C. Verghese (MIT), and Prof. Sandip Roy (Washington State) for many helpful discussions.